\documentclass[UTF8]{article}
\pdfoutput=1
\usepackage{hyperref}
\usepackage[T1]{fontenc}
\usepackage[utf8]{inputenc}
\usepackage{authblk}
\usepackage{amsmath}
\usepackage{amsfonts}
\newtheorem{theorem}{Theorem}

\newtheorem{defi}{Definition}

\usepackage{graphicx}
\usepackage{subfigure} 
\usepackage{float} 
\usepackage{subfigure} 
\usepackage{algorithm}
\usepackage[noend]{algpseudocode}
\usepackage{tikz,xcolor,hyperref}
\usepackage{tikz}
\usepackage{appendix}
\definecolor{lime}{HTML}{A6CE39}
\DeclareRobustCommand{\orcidicon}{
	\begin{tikzpicture}
	\draw[lime, fill=lime] (0,0) 
	circle [radius=0.16] 
	node[white] {{\fontfamily{qag}\selectfont \tiny ID}};
	\draw[white, fill=white] (-0.0625,0.095) 
	circle [radius=0.007];
	\end{tikzpicture}
	\hspace{-2mm}
}
\foreach \x in {A, ..., Z}{\expandafter\xdef\csname orcid\x\endcsname{\noexpand\href{https://orcid.org/\csname orcidauthor\x\endcsname}
		{\noexpand\orcidicon}}
}

\usetikzlibrary{graphs}
\usetikzlibrary{shapes.geometric,fit,positioning,arrows,automata,calc}
\tikzset{
	main/.style={circle, minimum size = 5mm, thick, draw =black!80, node distance = 10mm},
	connect/.style={-latex, thick},
	box/.style={rectangle, draw=black!100}
}
\title{Yet Another Representation  of Binary Decision Trees: A Mathematical Demonstration}
\author{Jinxiong Zhang \orcidA{}\\
jinxiongzhang@qq.com}
\date{} 
\begin{document}

\maketitle

\begin{abstract}

A decision tree looks like a simple directed acyclic computational graph, 
where only the leaf nodes specify the output values and the non-terminals specify their tests or split conditions. 
From the numerical perspective, we express decision trees in the language of computational graph. 
We explicitly parameterize the test phase, traversal phase and prediction phase of decision trees based on the bitvectors of non-terminal nodes. 
As shown, the decision tree is a shallow binary network in some sense. 
Especially, we introduce the bitvector matrix to implement the tree traversal in numerical approach, 
where the core is to convert the logical `AND' operation to arithmetic operations. 
And we apply this numerical representation to extend and unify diverse decision trees in concept.

\end{abstract}

\section{Introduction}

\subsection{Motivation}

Decision trees have been popular in diverse fields  as shown in
\cite{breiman2017classification, Maindonald2011Recursive, criminisi2012decision, criminisi2013decision, holzinger2015data}.
They are usually represented in two common types.
The first is the graphical representation in hierarchical or nested structure  similar to the data structure tree.
The second is to regard the decision tree as a collection of logical expression, 
where each expression corresponds to a pattern.
It is so-called logical representation.

Both representation types are originally designed for tabular data, where the data are organized like matrix.
The row of this kind matrix corresponds to one entity of interest;
the column corresponds to certain feature or attribute of the entities.

In deep neural networks, non-terminal nodes apply a nonlinear activation function (such as sigmoid or step function) to the input sample
and then they send them to the next layer until the last layer.
The inputs are transmitted layer by layer and share the same depth and activation function  during prediction.
The parameters are optimized based on the gradients of cost function during training. 
More complex architecture can include some `feedback structure'.

In decision trees, non-terminal nodes apply a Boolean test function to the input sample, i.e.,
they ask a `yes-or-no' question, such as the question `Is the first attribute is larger than 1?'.
Then they lead the input sample to the next node based on the answer until reaching the terminal node.
Different inputs have different depths and test functions in decision trees.
The test function depends on the `splitting criteria' to optimize during the growth of decision trees.
In vanilla binary decision tree, the test function does not change the inputs
and terminal/leaf node takes the mode of its instances ground truth empirical distribution as its prediction.

It seems that these two fields look so different from each other that there is few connection between them.
Are tree-based methods competitive rivals of deep learning methods?
How can we construct a counterpart of deep learning methods based on decision trees?
What can we do to unify or implement the tree-based methods and deep learning methods in a computational framework? 
We would like provide some mathematical theoretical clues to these problems.

\subsection{Related Work and Contribution}

The related topic with our attention is the integration of decision trees and deep learning.
There are some schemes to leverage neural networks and decision trees 
such as \cite{EntropyNet, kontschieder2015deep,wyner2017explaining, zhou2017deep, yang2018deep,hehn2019end, popov2019neural, hazimeh2020tree}.

Analogous to the fact that network and tree are two special types of graph,
we will show that decision trees and artificial neural networks are special types of computational graph.

Based on the bitvectors of internal nodes, we express decision trees as a numerical model.
The bitvector is originally adopted to improve the efficiency of the tree-based ranking model.
It is first time to illustrate how to perform an interleaved traversal of the ensemble
by means of simple logical bitwise operations according to
 \href{https://www.cse.cuhk.edu.hk/irwin.king/_media/presentations/sigir15bestpaperslides.pdf}{the slides}
presented in SIGIR15 by  Raffaele Perego for more details of QuickScorer.

As shown, this traversal is of fewer operations and insensitive to nodes' processing order.
It is the fewer operations that improves the practical performance of tree-based model.
Our motivation of the representation \eqref{tree} is to describe the decision tree model
in the language of computational graph
so that we can re-use deep learning software packages to implement tree-based models.
As a result, we must take the processing order of the node into consideration.
We convert the logical bitwise operations to the matrix-vector multiplication for tree traversal.
And we find  the false nodes via direct mathematical operations in our novel representation.
It is end-to-end as deep  neural network.
What is more, our representation is compatible with diverse decision trees as shown later.
In theory, we introduce and generalize the bitvector matrix.
And we find that the bitvector matrix  determines the structure of decision tree.

In \cite{fridedman1991multivariate}, the recursive partitioning regression, as binary regression tree,  
is  viewed in a more conventional light as a stepwise regression procedure:
\begin{equation}\label{sum-product}
f_M(\mathbf{x})=\sum_{m=1}^{M}a_m\prod_{k=1}^{K_m}H[s_{km}(x_{v(k, m) - t_{km}})]
\end{equation}
where $H(\cdot)$ is the unit step function.
The quantity $K_m$, is the number of splits that gave rise to basis function.
The quantities $s_{km}$ take on values k1and indicate the (right/left) sense of the associated step function.
The $v(k, m )$ label the predictor variables and 
the $t_{km}$, represent values on the corresponding variables.
The internal nodes of the binary tree represent the step functions and the terminal nodes represent the final basis functions.
It is hierarchical model using a set of basis functions and stepwise selection.
The product term $\prod_{k=1}^{K_m}H[s_{km}(x_{v(k, m) - t_{km}})]$ is an explicit form of the indicator function $\mathbb{I}$.  
In \cite{yang2018deep}
soft binning function is used to make the split decisions.
Typically, a binning function takes as input a real scalar $x$
and produces an index of the bins to which $x$ belongs.
Yongxin Yang et al \cite{yang2018deep} propose  deep neural decision tree as a differentiable approximation of this function.
Given the differentiable approximate binning function,
it is to construct the decision tree via Kronecker product $\otimes$.
They asserted that they can exhaustively find all final nodes by
$$f_1(x_1)\otimes f_2(x_2)\otimes\cdots\otimes f_D(x_D)$$
where each feature $x_d$ is binned  by its own neural network $f_d(x_d)$.
And linear classifier is added as the last layer of deep neural decision tree in \cite{yang2018deep}.
That work holds the properties we desire:
the model can be easily implemented in neural networks tool-kits,
and trained with gradient descent rather than greedy splitting.

However, our representation is to describe the decision trees in the computational graph language precisely.
and it is proved to be equivalent to the common representation of decision trees based on a solid theoretical foundation.
It is not hybrid or analogous of deep neural network, which makes it unique within diverse blend scheme of decision trees and deep learning models.
And our representation can deal with categorical feature.
In short, we intrinsically reshapes the characteristics of the decision trees from the analytic aspects.

Our main conceptional contributions includes:
\begin{enumerate}
	\item a compact form which explicitly parameterize the decision trees in the terms of computational graph;
	\item a configurable framework which unifies and extends diverse tree-based methods;
	\item a bridge which connects the deep learning and decision trees.
\end{enumerate}

In the following sections, a new representation of decision tree is proposed in the language of computational graph.
The decision tree is shown to be a shallow binary network in some sense.
In another word we extend the field of computational graph beyond deep learning
and acceleration techniques in deep learning may help to accelerate the tree-based learning algorithms.
A specific example is to illustrate this representation.
More characters of this representation are discussed in the section on generalization of decision trees.
And we apply such new parameterization of decision trees to unify some hybrid decision trees.

\section{Parameterized Decision Trees}

\subsection{A Parameterization of Decision Trees}

Different from building a decision tree, prediction of a decision tree is the tree traversal in essence.
Inspired by \href{http://pages.di.unipi.it/rossano/wp-content/uploads/sites/7/2015/11/sigir15.pdf}{QuickScorer}\cite{lucchese2015quickscorer},
we split such prediction to the following steps.
We only consider the numerical features for simplicity here and categorical features are discussed later.

The first step is to  find the false nodes in the decision tree with respect to the input $\mathbf{x}\in\mathbb{R}^n$:
   \begin{equation}\mathbf{h}=\frac{\operatorname{Sgn}(\mathrm{S}\mathbf{x}-\mathbf{t})+\vec{1}}{2}=\sigma(\mathrm{S}\mathbf{x}-\mathbf{t})\end{equation}
where so-called `selection matrix' $S\in\mathbb{R}^{n_L\times n}$ consists of one-hot row vector in $\mathbb{R}^n$
representing which feature is tested in a node;
the elements of threshold vector $\mathbf{t}\in\mathbb{R}^{n_L}$ are the optimal splitting points of each node associated with one feature;
$\operatorname{Sgn}(\cdot)$ is the element-wise sign function and $\operatorname{Sgn}(0)=-1$; 
$\sigma(x)$ is the \href{https://arxiv.org/pdf/1901.09731.pdf}{binarized ReLU}\cite{dinh2019convergence} function defined by
$$
\sigma(z)=\begin{cases}1, &\text{if $z> 0$}\\
0, &\text{otherwise}\end{cases}.
$$
Here $n_L$ is the number of the non-terminals and $n$ is the dimension of input space.
For convenience, we define that the node is a true node if its feature is greater than the splitting point
otherwise the node is `false' node.
We call this step as \emph{test phase}.

The second step is to find the index corresponding to the exit leaf based on the bitvector\footnotemark matrix of the decision tree
\footnotetext{Every non-terminal node $n$ is associated with
	a node \href{http://pages.di.unipi.it/rossano/wp-content/uploads/sites/7/2015/11/sigir15.pdf}{bitvector}
	(of the same length), acting as a bitmask that encodes (with 0’s)
	the set of leaves to be removed from the subtree whenever $n$ is a false node.
	It is declared that \href{https://dl.acm.org/citation.cfm?id=2987380}{Domenico Dato et al}
	are the first to represent the non-terminal nodes as bitvectors.}
\begin{equation}
	\mathbf{p} = \mathrm{B}\mathbf{b}+\vec 1,\quad
	i = \arg\max(\mathbf{p})
\end{equation}
where 
the matrix $\mathrm{B}\in\mathbb{B}^{L\times n_L}$ is the bitvector matrix of the decision tree; 
the integer $n_L$ is the number of the non-terminal nodes; 
$\arg\max(\mathbf{p})$\footnotemark returns the smallest index of maximums in the vector $p$,
i.e., $i=\arg\max(\mathbf{p})=\min\{j\mid p_j=\max(\mathbf{p})\}$.
Every column of $\mathrm{B}$ is a bit-vector of node; the matrix $\mathrm{B}\mathbf{b}$
is the matrix multiplication product of matrix $\mathrm{B}$  and $\mathbf{h}$.
Here $L$ is the number of terminal nodes(leaves).
And the columns of true nodes will be multiplied by 0es and the columns of false nodes will keep unchanged.
We call this step \emph{traversal phase}.
\footnotetext{Note that the operator $\arg\max(\mathbf{p})$ generally returns all indices of the max element of the array $p$, i.e.,  
$\arg\max(\mathbf{p})=\{j\mid p_j=\max(\mathbf{p})\}$.}

The final step is to determine the output
\begin{equation}v[i],\end{equation}
where 
$v[i]$ are the $i$th elements of vector $v$.
We call this step \emph{prediction phase}.

If $h=\vec 0$, we will reach the  leftmost node in the decision tree and if $h=\vec 1$, we will reach the rightmost node in the decision tree.
Thus it is sensitive to the order of leaves. And we assume that nodes  are numbered in {\color{red} breadth-first order and leaves from left to right}
as in \cite{lucchese2015quickscorer}.

\begin{algorithm}
	\caption{\textit{QuickScorer} in language of matrix computation}\label{alg:1}
	\begin{algorithmic}[1]
		
		\Statex \textbf{Input}:
		\begin{itemize}
			\item  input feature vector $\mathbf{x}$
			\item  a binary decision tree  with $v, \mathrm{B}, \mathrm{S}, \mathbf{t}$

		\end{itemize}
		
		\Statex \textbf{Output}: tree traversal output value
		\Procedure{Scorer}{$\mathbf{x}, v, \mathrm{B}, \mathrm{S}, \mathbf{t}$} 
		\State Compute the latent test vector $\mathbf{h}=\frac{\operatorname{Sgn}(\mathrm{S}\mathbf{x}-\mathbf{t})+\vec{1}}{2}$. \Comment{test phase}
		\State  $\mathbf{p}= \mathrm{B}\mathbf{b}+\vec 1$. 
		\State $j\leftarrow$ the smallest index of the maximum of $\mathbf{p}$. \Comment{Traversal phase}
		\State return $v[j]$.  \Comment{Prediction phase}
		
		\EndProcedure
		
	\end{algorithmic}
\end{algorithm}

Note that the key is the index of the maximums in the  hidden state
$$[\mathrm{B}\underbrace{\sigma(\mathrm{S}\mathbf{x}-\mathbf{y})}_{\text{binary vector}}+\vec 1]$$
where if the test returns true, the  element of the binary vector is zero;
otherwise, the element of the binary vector is 1.
The 1s select the bitvectors of false nodes and
the 0s rule out the bitvectors of true nodes.
If we change $1$ to any positive number,
the first index of the maximums in the hidden state does not change.
Thus, binarized ReLU is not the unique option of activation function.
For example, \href{https://arxiv.org/abs/1611.01491}{ReLU} is an alternative to binarized ReLU in our representation.
We will show more alternatives of binarized ReLU later.
We can rewrite the decision tree in a more compact form:
\begin{equation}\label{tree}
T(\mathbf{x})=v[\arg\max([\mathrm{B}\underbrace{\sigma(\mathrm{S}\mathbf{x}-\mathbf{y})}_{\text{binary vector}}+\vec 1])]
\end{equation}
where the notations are defined as above.
And the consatnt vectors $p$ and $\vec 1$ are used to ensure the maximum of the result is unique in (\ref{tree}), 
which are omitted based on our irregular $\arg\max(\cdot)$ operator definition in the rest of this paper.

Here is the data flow  of decision tree in this case:
\begin{equation}\label{hidden states}
\underbrace{\mathbf{x}}_{\mathbb{R}^n} 
\xrightarrow{\mathrm{S}\mathbf{x}-\mathbf{y}} \underbrace{\tilde{h}}_{\mathbb{R}^{n_L}} 
\xrightarrow{\sigma(\tilde{\mathbf{h}})} \underbrace{\mathbf{h}}_{\mathbb{B}^{n_L}} 
\xrightarrow{\mathrm{B}\mathbf{b}} \underbrace{\mathbf{p}}_{\mathbb{Z}^{L}} 
\xrightarrow{i=\arg\max(\mathbf{b})} \underbrace{v[i]}_{\mathbb{O}} \nonumber
\end{equation}

Here $\mathbb{B}^{n_L}$ is the Boolean space $\{0, 1\}^{n_L}$;
$\mathbb{R}^n$ is n-dimensional real space;
$\mathbb{Z}^{n_L}$ is Cartesian product of integer set.
In regression, $\mathbb{O}$ is the Euclidean space $\mathbb{R}$.
Additionally, the element of value vector $v$ is categorical for classification trees.

It is really a shallow model as the above relation\eqref{hidden states} shown.
We can observer that there is  a single hidden layer with the binarized ReLU function in the representation\eqref{tree}.
This would help us understand the binary neural networks further
if we can understand the inherent connection the decision trees and neural networks further.
In contrast to the sequential tests of vanilla decision trees, we will take all the tests in parallel even simultaneously at \eqref{tree}.

Now there is nothing other than expressing the decision tree in the language of computational graph.
It looks  far from a step function. However, note that
\begin{itemize}
\item the matrices $\mathrm{S}$ and $\mathrm{B}$ are binary, i.e., their elements are 0 or 1.
\item $\arg\max()$ only selects one element in this case.
\end{itemize}
All `if-then' sentences are transformed to numerical computation.
And the test phase and the traversal phase are split and separated.
In test phase, we obtain the hidden state $\mathbf{h}$ dependent on the selection matrix $\mathrm{S}$ and threshold vector $\mathbf{t}$: 
$\mathbf{h}=\sigma(\mathrm{S}\mathbf{x}-\mathbf{b})$.
In traversal phase, we get the hidden state $\mathbf{p}$ dependent on the bitvector matrix $\mathrm{B}$: $\mathbf{p}=\mathrm{B}\mathbf{b}$
and then arrive at the terminal node by finding the first index of the  maximums in  the product: $v[i], i=\min\{j\mid l_j=\max(l)\}$.
These  phases play different roles while they are connected and associated.
All the input samples share the same computational procedure.
This makes it more computation intensive  in prediction.
The good news is that we can apply the high performance techniques such as fast matrix multiplication to decision trees.

The tuple $(v, \mathrm{B}, \mathrm{S}, \mathbf{t})$ may determine a decision tree and we propose the following statement:
\begin{center}
	\noindent\fbox{
		\parbox{0.9\linewidth}{	
		When the processing order of the nodes are given, there is an one-to-one $M: T(n) \mapsto (v, \mathrm{B}, \mathrm{S}, \mathbf{t})$
		where  $T(n)$ is the set of  all  the decision trees where the number of non-terminal nodes is $n$; 
		the tuple $(v, \mathrm{B}, \mathrm{S}, \mathbf{t})$ is defined as above.}
	}
\end{center}

And the formula \eqref{tree} is so-called the third representation of decision tree, parameterization representation.
We provide some evidence of above statement  and some theoretical properties of the bitvector matrix in the appendix.

Based the this representation tuple $v, \mathrm{B}, \mathrm{S}, \mathbf{t}$ of decision tree, we can identify every binary decision tree.
All tree-based algorithms will benefit from such representation during prediction rather than training.

\subsection{An Example of Binary Decision Trees}

Suppose the input variable $x$ has 4 numerical features,
the learned decision tree is as shown in the following figure \ref{Fig.main1}.

The threshold vector is $\mathbf{t}=(1,4,3,2,5)^T$.
The value vector is  categorical or numerical,
such as $\mathbf{v}=(v_1, v_2, v_3, v_4, v_5, v_6)^T$.
The selection matrix $\mathrm{S}$ and  bit-vector matrix $\mathrm{B}$ is
$$
\begin{pmatrix}
1 & 0 & 0 & 0\\
0 & 1 & 0 & 0\\
0 & 0 & 1 & 0\\
0 & 1 & 0 & 0\\
0 & 0 & 0 & 1
\end{pmatrix},
\begin{pmatrix}
0 & 0 & 1 & 0 & 1\\
0 & 0 & 1 & 1 & 0\\
0 & 0 & 1 & 1 & 1\\
0 & 1 & 1 & 1 & 1\\
1 & 1 & 0 & 1 & 1\\
1 & 1 & 1 & 1 & 1\\
\end{pmatrix},
$$
respectively.
And $\sigma$ is the binarized ReLU.

\begin{figure}[H] 
\centering 
\tikzstyle{results}=[ellipse ,text centered,draw=black]
\tikzstyle{decisions} =[rectangle, rounded corners,text centered, draw = black]
\tikzstyle{arrow} = [-,>=stealth]

\begin{tikzpicture}[node distance=0.8cm]
\node[decisions](rootnode){ $\mathbf{x}[1]\leq 1?$ };
\node[decisions,below of =rootnode,yshift=-0.3cm,xshift=-2.0cm](point1){$\mathbf{x}[2]\leq 4?$};
\node[decisions,below of =rootnode,yshift=-0.3cm,xshift=2.0cm](point2){$\mathbf{x}[3]\leq 3?$};
\node[decisions,below of =point1,yshift=-0.3cm,xshift=-1.2cm](point3){$\mathbf{x}[2]\leq 2?$};
\node[decisions,below of =point1,yshift=-0.3cm,xshift=1.2cm](leaf){$v_4$};
\node[decisions,below of =point2,yshift=-0.3cm,xshift=-1.2cm](leaf1){$v_5$};
\node[decisions,below of =point2,yshift=-0.3cm,xshift=1.2cm](leaf2){$v_6$};
\node[decisions,below of =point3,yshift=-0.3cm,xshift=-1.2cm](leaf3){$v_1$};
\node[decisions,below of =point3,yshift=-0.3cm,xshift=1.2cm](point4){$\mathbf{x}[4]\leq 5?$};
\node[decisions,below of =point4,yshift=-0.3cm,xshift=-1.2cm](leaf4){$v_2$};
\node[decisions,below of =point4,yshift=-0.3cm,xshift=1.2cm](leaf5){$v_3$};
\draw [arrow] (rootnode) -- node [left,font=\small] {True} (point1);
\draw [arrow] (rootnode) -- node [right,font=\small] {False} (point2);
\draw [arrow] (point1) -- node [left,font=\small] {True} (point3);
\draw [arrow] (point1) -- node [right,font=\small] {False} (leaf);
\draw [arrow] (point2) -- node [left,font=\small] {True} (leaf1);
\draw [arrow] (point2) -- node [right,font=\small] {False} (leaf2);
\draw [arrow] (point3) -- node [left,font=\small] {True} (leaf3);
\draw [arrow] (point3) -- node [right,font=\small] {False} (point4);
\draw [arrow] (point4) -- node [left,font=\small] {True} (leaf4);
\draw [arrow] (point4) -- node [right,font=\small] {False} (leaf5);
\end{tikzpicture}
\caption{A Binary Decision Tree} 
\label{Fig.main1} 
\end{figure}
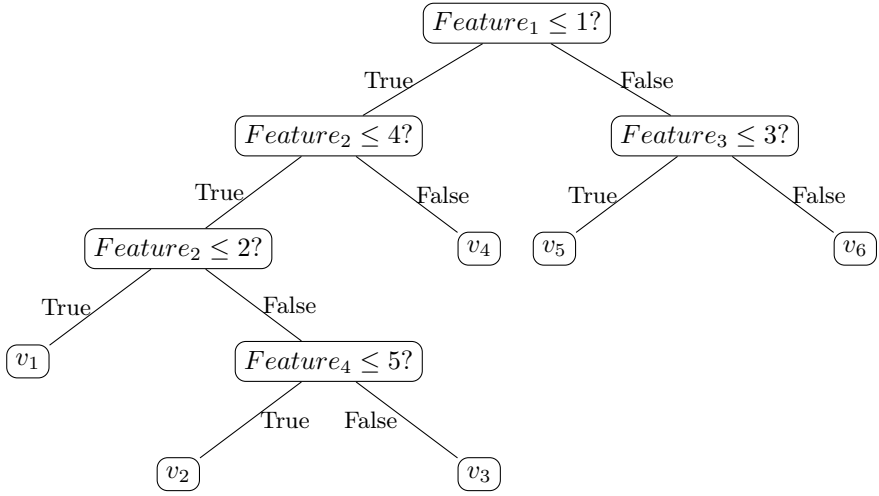

Suppose input vector $\mathbf{x}=(2,1,2,2)^T$, its prediction is $v_5$ according to the figure \ref{Fig.main1}.
The computation procedure is as following
$$\tilde h=\mathrm{S}\mathbf{x}-\mathbf{t}=\begin{pmatrix}
1\\ -3\\ -1\\ -1\\ -3 
\end{pmatrix},
\mathbf{h}=\sigma(\tilde h)=\begin{pmatrix}
1\\ 0\\ 0\\ 0\\ 0 
\end{pmatrix},
\mathbf{p}=\mathrm{B}\sigma(\mathbf{h})=\begin{pmatrix}
0\\ 0\\ 0\\ 0\\ 1\\1 
\end{pmatrix},$$
$$i=\arg\max(\mathbf{p})=5,v[i]=v_5,$$ 
thus then the output is given by $T(\mathbf{x})=v_5$.

Here we do not specify the data types of the elements in the value vector.
For regression, the value vector $v$ is numerical;
for classification, it is categorical \footnotemark.
\footnotetext{See more differences of classification trees and regression trees in
\href{http://pages.stat.wisc.edu/~loh/treeprogs/guide/wires11.pdf}{Classification and regression trees}.}

The first element of bitvector always corresponds to the leftmost leaf in the decision tree
so that the algorithm return the first value when there is no false node for some inputs such as $(1,1,2, 3)^T$.
And the order of bitvector is left to right, top to bottom.

\subsection{Categorical Features}

It is always to take different measures to deal with qualitative and quantitative features.
For example, \href{https://cs.nju.edu.cn/zhouzh/zhouzh.files/publication/kbs02.pdf}{Hybrid Decision Tree}\cite{zhou2002hybrid}
simulates human reasoning by using
symbolic learning to do qualitative analysis and using neural learning to do subsequent quantitative analysis.
\href{https://arxiv.org/abs/1604.06737}{Cheng Guo and Felix Berkhahn}\cite{guo2016entity}
embed the categorical features to Euclidean spaces.
\href{http://learningsys.org/nips17/assets/papers/paper_11.pdf}{CatBoost}\cite{prokhorenkova2017catboost:},
a new open-sourced gradient boosting library, successfully handles categorical features.
\href{http://www.jmlr.org/papers/volume18/17-716/17-716.pdf}{Elaine Angelino, Nicholas Larus-Stone and Daniel Alabi et al}\cite{angelino2017learning} 
invented a novel alternative to CART and other decision tree methods,
which uses a custom discrete optimization technique for building rule lists over a categorical feature space.

Categorical features are usually stored in different data format from the numerical features in computer,
which can provide some qualitative information for prediction.
The domain/value set of categorical feature is discrete and usually finite.
The categorical features are often unordered and sometimes ordinal. 
For example, we cannot compare the gender in the magnitude and 
we would not say the your gender is `larger or smaller' than my gender.
Note that there are no arithmetic operations of categorical features.
The only operation we can perform on categorical features is to test whether it is of one specific type.
For example, we may ask the question ``Is the color red?'' to classify goode apples.
That is a `yes-no' question and we would take the next step according to the answer.

Because we cannot add up the numerical features and categorical features directly,
we should split embedding representation of categorical features and numerical features in oblique decision tree
while there is no problem  to test categorical or numerical features in vanilla decision tree \footnotemark.

\footnotetext{The combination of categorical features are discussed in
\href{http://learningsys.org/nips17/assets/papers/paper_11.pdf}{CatBoost}.
However, it is Cartesian product of some domains rather than linear combination of different features.}

After embedding the categorical features into real vectors,
we can express the decision trees with categorical feature in the following formula:
$$T(\mathbf{x})=v[\arg\max(\mathrm{B}\,\sigma(\mathrm{S}\mathbf{x}-\mathbf{y}))]$$
where $\mathrm{S}$ is the one-hot row vector  matrix and
$\mathrm{B}$ is the bitvector column matrix.
The threshold value of categorical feature is also limited in the embedded (discrete) space.
For example, if the categorical variables are stored in the format of string in computer
and we use the hash function to transform them to integers as embedding,
then the embedded integers are legal for the axis-aligned mapping $\mathrm{S}\mathbf{x}$.
Inspired by \cite{O'Sullivan1991multivariate}, we can the Lagrange polynomials. 

Now we define the oblique decision trees with categorical features.
As mentioned above, it is not possible to add up the numerical features and categorical features directly.
For example, there is no sense to add your age and gender.
They introduce the dummy variables(sometimes called indicator variable) in statistics and quantitative economy community
such as \href{https://www.sagepub.com/sites/default/files/upm-binaries/21120_Chapter_7.pdf}{Dummy-Variable Regression}.
\href{http://www.economicswiki.com/economics-tutorials/dummy-variable/}{Economics Wiki} explains that
``This indicator variable takes on the value of 1 or 0 to indicate the availability or lack of some effect
that would change the outcome of whatever is being tested."
If we regard the linear combination of numerical features as classic linear regression
to test some properties of input sample,
we extend the linear regression to dummy-variable regression or regression discontinuity design(RDD).
Thus we can deal with the categorical features in oblique decision trees without embedding:
\begin{equation}\label{DTC}
T(\mathbf{x}) =v[\arg\max(\mathrm{B}\,\sigma(f(\mathbf{x})))] \\ \nonumber
\end{equation}
where $f(\mathbf{x})=(f_1(\mathbf{x}), f_2(\mathbf{x}), \cdots, f_{n_L}(\mathbf{x}))^T$ and $f_{i}(\mathbf{x})$ is the dummy-variable regression.

\section{Application}

The new parameterization formula of decision trees are applied to generalize and unify diverse decision trees.
Additionally, any tree-based algorithm can benefit from this new parameterization.

\subsection{Extension of Binary Decision Trees}

In this section, we will show how to extend each component of the parameterized decision trees\eqref{tree}.

\subsubsection{The Feature Transformation Mapping}

It is the matrix $\mathrm{S}$ which is the difference of vanilla decision trees and the oblique decision trees. 
It is one-hot vector in the column space of selection matrices $\mathrm{S}$ associated with vanilla decision trees 
while it is real column vector in the selection matrices $\mathrm{S}$ of oblique decision trees.
Oblivious decision trees specify the selection matrices $\mathrm{S}$ if the number of leaf node is given.
Our framework can include more hybrid decision trees.
For example, \href{https://erwanscornet.github.io/pdf/articlekernel.pdf}{Erwan Scornet},
\href{https://arxiv.org/pdf/1402.4293.pdf}{Alex Davies and Zoubin Ghahramani} apply kernel tricks to tree-based algorithms.
We can replace the univariate or multivariate linear functions $\mathrm{S}\mathbf{x}-\mathbf{t}$ with nonlinear transformation functions $f=(f_1,\cdots,f_{n_L})^T$:
\begin{equation}\label{GT1}
  T(\mathbf{x})=v[\arg\max(\mathrm{B}\,\sigma(f(\mathbf{x})))].
\end{equation}
In other words, it is to project the raw input samples $x$ to the range of $f$.
The mapping $f:\mathbb{R}^n\mapsto \mathbb{R}^{n_L}$ is used to extract common features and explore the underlying patterns of raw inputs
so we call it  `feature transformation mapping' and its component $f_i$ as `feature transformation function'.

As other supervised learning, the choice of feature transformation function is an important part of feature engineering.
The mapping $\sigma(f(\mathbf{x}))$ reflects our tests on the raw input sample thus $f$ 
is to digitalize our prior experience and knowledge on the task to learn in some sense.

For example, \href{https://www.ijcai.org/Proceedings/2019/0493.pdf}{Philip Tannor and Lior Rokach} apply the features from the last hidden
layer (supervised) of neural networks to augment the features in the dataset between the iterations of Gradient Boosted Decision Trees(GBDT).
In our language, they use a trained neural network as the feature transformation mapping in each decision tree of the GBDT.
It is  the modification of feature transformation mapping that is significant for some variants of decision trees for so-called manifold-valued objects such as 
\href{http://citeseerx.ist.psu.edu/viewdoc/download?doi=10.1.1.676.9232&rep=rep1&type=pdf}{Random Forest Manifold Alignment},
\href{https://www.ncbi.nlm.nih.gov/pubmed/18779085}{Decision Manifolds},
\href{https://arxiv.org/abs/1909.11799}{Manifold Forests},
\href{https://neurodata.io/sporf/}{SPORF (Sparse Projection Oblique Randomer Forests)},
\href{https://labicvl.github.io/docs/pubs/Bonde_ACCV_2010.pdf}{Randomised Manifold Forests}.

It reveals the inherent dual nature of decision trees:
on one hand, it is rule-based because the type and the number of test functions are pre-specified  like a questionnaire;
in another hand, it is data-based because the test functions are updated dynamically according to the statistical information of input samples. 

\subsubsection{The Activation Function}

There are many activation functions in deep neural network such as
ReLU\cite{dinh2019convergence},
\href{https://arxiv.org/pdf/1302.4389.pdf}{maxout}
\href{https://arxiv.org/pdf/1502.01852.pdf}{parametric ReLU},
\href{https://papers.nips.cc/paper/6698-self-normalizing-neural-networks.pdf}{SELU},
\href{https://arxiv.org/pdf/1710.05941.pdf}{Swish},
\href{https://arxiv.org/abs/1908.08681}{Mish}.
There are a partial list of activation functions in deep neural network at
\href{https://www.simonwenkel.com/2018/05/15/activation-functions-for-neural-networks.html}{Simon's blog}
and a more comprehensive list of activation functions
at \href{https://en.wikipedia.org/wiki/Activation_function}{Wikipedia}.
They are motivated initially to solve the vanishing gradient problem
and boost the stability as well as the generalization ability.
In contrast to deep learning, there is no worry about the gradient vanishing or exploding problem during decision tree growth. 
However, we cannot apply  activation functions in deep learning to decision trees without verification.
It is not hard to verify that ReLU is an eligible alternative to the binarized ReLU or unit step function in \eqref{tree}
as well as its variants $\tau(z)=\alpha\max \{0, z\}$(so-called positively scaled ReLU) and 
$\tau(z)=\max\{0, \operatorname{sgn}(z)z^2\}$(so-called rectified quadratic unit).

In fact, any positivity-augmented function $\tau$ is an alternative to the binarized ReLU or unit step function in \eqref{tree}, i.e.,
$$v[\arg\max(\mathrm{B}\sigma(\mathrm{S}\mathbf{x}-\mathbf{t}))]=v[\arg\max(\mathrm{B}\tau(\mathrm{S}\mathbf{x}-\mathbf{t}))].$$
\begin{defi}
The $\tau(z):\mathbb{R}\mapsto\mathbb{R}^{+}$ is positivity-augmented 
if it is positive when the input $z$ is positive otherwise  0.
\end{defi}
It is used to augment the positive evidence or signal of the  tests and rule out the false or negative samples.
It maps the non-positive part of real number $R$ to the original point so it is a truncation type function. 
For example, the cumulative distribution function $\sigma$ is an positivity-augmented function
if its support is $(0,\infty)$ so that if $z\leq 0$ then $\sigma(z)=0$;
otherwise $\sigma(z)>0$ and $\lim_{z\to\infty}\sigma(z)=1$.
It is easy to adapt the  penalty function such as \href{https://projecteuclid.org/download/pdfview_1/euclid.aos/1266586618}{MC+} for variable selection 
into positivity-augmented function.
The penalty function $p$ is usually symmetric and non-negative, i.e.,
\begin{itemize}
  \item  $p(z)=p(-z)$ and $p(0)=0$;
  \item  $p(z)\geq 0$. 
\end{itemize}
If $p(\cdot)$ holds both properties, it is simply to adapt $p$ to a positivity-augmented function via 
$p^+=\max\{0, \operatorname{sgn}(x) p(x)\}$
where $\operatorname{sgn}$ is the sign function. 
It is also possible to modify odd functions to positivity-augmented functions such as $\tau(z)=\max\{0, z+\sin(z)\}$ and $\tau(z)=max\{0, z^3\}$.
In fact, we can construct more general positivity-augmented functions such as $\tau(z)=\max\{0,\min\{z, \frac{1}{z}\}\}$, $\tau(z)=\max\{0, \frac{1}{|\log(z)|}\}$.

The positivity-augmented functions are often piecewise but that is not necessary.
These functions are not required to be monotonic or differentiable.
In theory, we can apply different activation functions to different test functions, 
i.e, the equation\eqref{tree} can be modified as following
$$T(\mathbf{x})=v[\arg\max(\mathrm{B}\mathbf{b})],\\ h_i=\tau_i([(\mathrm{S}\mathbf{x}-\mathbf{t})]_i)$$
where $[(\mathrm{S}\mathbf{x}-\mathbf{t})]_i$ is the $i$th element of the vector $(\mathrm{S}\mathbf{x}-\mathbf{t})$.

\subsubsection{The Bitvector Matrix}

The bitvector matrix $\mathrm{B}$ in \eqref{tree} is Boolean and friendly to encode thus it saves the storage space and boots the performance.
In another hand, it is hard-encoded and unfriendly to update dynamically.
If we want to train a decision tree\eqref{tree} from scratch, it requires a more convenient code scheme of the bitvector matrix.
Like the \href{https://web.ma.utexas.edu/users/gordanz/notes/lebesgue_integration.pdf}{Standard Machine} in Lebesgue Integral,
we generalize the bitvector matrices step by step.
 
Similar to the extension of activation function, we will obtain the first rule:
it does not change the first index of the maximum in $B\sigma(\mathrm{S}\mathbf{x}-\mathbf{t})$ for the decision tree \eqref{tree} 
if we only replace the number $1$ in the matrix $\mathrm{B}$ with the identical positive constants.
Observe that $\mathrm{B}\mathbf{b}=\sum_{i=1}^{n_L}h_i B_i$, 
we can positively scale $B_i$ with $\alpha_i > 0$,
i.e., $\arg\max(\mathrm{B}\mathbf{b})=\arg\max(B\operatorname{diag}(\alpha)h)$,
where $h\in\mathbb{B}^{n_L}$ and $B\in\mathbb{R}_{+}^{L\times n_L}$ is nonnegative matrix.
In another hand, it is equivalent to positively scale $h_i$ with $\alpha_i > 0$.

And when $\arg\max(\mathrm{B}\mathbf{b})=\arg\max(BMh)$, what is the sufficient and necessary condition of the matrix $M$?
The matrix $M$ must preserve the first index of maximum in the non-negative vector.

The second rule is to substitute any negative constant  for each zero while keep others unchanged over the matrix $\mathrm{B}$ in \eqref{tree}, 
Combining both rules, we would replace the ones with identical positive constants  and the zeroes with the negative constants.
For example, we can replace the matrix $\mathrm{B}$ in \ref{Fig.main1} with following matrix:
$$\begin{pmatrix}-2 & -200& 2 &-3 & 1\\ -3 & -400& 2 & 4 &-1\\ -4 & -4  & 2 & 4 & 1\\ -5 & 20 & 2 & 4 & 1 
\\ 5.1& 20 &-2 & 4 & 1 \\ 5.1& 20 & 2 & 4 & 1\\ \end{pmatrix}.$$ 
It is compatible with the logical `AND' operation of the bitvectors:
the  ones in the bitvectors are referred to the potential leaf nodes to reach
while the  zeroes are referred to the leaf nodes impossible to reach.  
The $0$ is the veto signal in logical `AND' operation.
When transferring from logical operation to numerical operation, 
the zeroes in the bitvector is ideally adapted to  $-\infty$ as a strong negative signal.
At least the zeroes are supposed to be transferred to  non-positive as a negative signal.

Any matrix $M$ can be expressed as the difference of two non-negative matrices $P_1-P_2$.
For example, 
$$\begin{pmatrix}-2 & -2  & 2 &-3 & 1\\ -3 & -1  & 2 & 4 &-1\\ -4 & -4  & 2 & 4 & 1\\ 
-5 & 6.2 & 2 & 4 & 1\\ 5.1& 6.2 &-2 & 4 & 1\\ 5.1& 6.2 & 2 & 4 & 1\\ \end{pmatrix}=
\begin{pmatrix}0  & 0   & 2 & 0 & 1\\ 0  & 0   & 2 & 4 & 0\\ 0  & 0   & 2 & 4 & 1\\
0  & 6.2 & 2 & 4 & 1\\ 5.1& 6.2 & 0 & 4 & 1\\ 5.1& 6.2 & 2 & 4 & 1\\ \end{pmatrix}
-\begin{pmatrix}2 & 2  & 0 & 3 & 0\\ 3 & 1  & 0 & 0 & 1\\ 4 & 4  & 0 & 0 & 0\\
5 & 0  & 0 & 0 & 0\\ 0 & 0  & 2 & 0 & 0\\ 0 & 0  & 0 & 0 & 0\\ \end{pmatrix}.$$ 
It is clear that the domains of the $P_1$ and $P_2$ are different. 
This lack of symmetry is because of  the logical `AND' operation
$$\mathrm{B}_1\wedge \mathrm{B}_2\wedge\cdots \mathrm{B}_m= \mathrm{B}_1\circ \mathrm{B}_2\circ\cdots \mathrm{B}_m$$
where $\wedge$ is logical `AND' and $\circ$ is element-wise multiplication.
We can observe that it is the number of $0$ that play the lead role of  the results of logical `AND' or numerical multiplication,
i.e., only one zero can lead to the result 0 of logical `AND' or numerical multiplication.
In contrast, the result is 1 if and only if all operands are 1.
The logical `AND' operation only process the yes or no.
What is worse, we cannot express the double negation with logical `AND'.
This is also compatible with case of the activation function extension
where we only enhance the affirmative answer.

Above schemes of extension focus on the column space of bitvector matrices and preserve the logical interpretation of bitvector matrix.

Now we extend the bitvector matrices to all 0-1 matrices.
We obtain the following generalized decision trees:
\begin{equation}\label{GT}
   T(\mathbf{x})=v[i], i=\arg\max{[\mathrm{B}\,\sigma(\mathrm{S}\mathbf{x}-\mathbf{t})+\vec 1]}
\end{equation}
where the matrix $\mathrm{B}$ is a Boolean/logical matrix, i.e., its element is $0$ or $1$;
other notations have the same meaning as in \eqref{tree}.
This extension makes some sense.
For example, if $\mathrm{B}$ is the matrix full of ones, the formula \eqref{GT} is a constant function $T(\mathbf{x})=v[1]=v[2]=\cdots=v[L]$.
However, there is no equivalent logical `AND' operation of the columns of the matrix $\mathrm{B}$ in above example. 

If the bitvector is extended to real vector space,
the vanilla decision tree\eqref{tree} is sparse in the new representation.
If we still keep the bitvector in Boolean space,
the classical decision tree is not too rare  in our new representation.
This extension includes the decision tree.
In another word, our new representation is the super set of decision tree.

In \eqref{GT}, the most general form of $\mathrm{B}$ is the one whose each element takes the real number as its domain. 
We call the formula\eqref{tree} `decision machine' if $\mathrm{B}$ is real
because most of them cannot be interpreted in the tree-traversal way as the classical original decision trees.
We turn to the row space of the bitvector matrix.
Each element of the result $\mathrm{B}\mathbf{b}$ is the inner product of the row vector of $\mathrm{B}$ such as $\mathrm{B}$ and the vector $\mathbf{h}$,
which measures the similarity of the vector $\mathrm{B}$ and $\mathbf{h}$ in some sense. 
Thus the result $\mathrm{B}\mathbf{b}$ can be interpreted as combination of multiple models,
where the largest one is selected to determine the output.
And in this case the activation function can extend to negative part of real number.
It is less logical, more numerical and closer to neural network.

Let us apply this interpretation to the vanilla decision tree.
The input sample follow a path from the root to the terminal/leaf node, which we call it  decision path to the leaf node.
\begin{theorem}\label{theorem_zhang}
Each decision path is equivalent to a sequence of test in decision tree.
\end{theorem}
We call the above theorem \emph{zeroth law of decision trees}. 
For example, if a sample arrives at the leaf node $v_3$ in \ref{Fig.main1}, 
it must traverse the first, the second, the fourth and the fifth non-terminal nodes
and it must pass the first and the second tests and fail the fourth and the fifth tests.
Additionally, it does not take the third test.
For the leaf node $v_3$, it is not necessary to know the results of the third test.
The third row of the corresponding bitvector matrix of \ref{Fig.main1} cannot reveal all the above information.
There are 3 types of signal for the terminal/leaf nodes as below
\begin{enumerate}
  \item $-1$ to represent the passed tests;
  \item $+1$ to represent the failed tests;
  \item $0$ to represent the absent tests.
\end{enumerate}
For example, it is better to use $(-1, -1, 0, +1, +1)$ to represent the leaf node $v_3$ in \ref{Fig.main1}.
And in this case, we can apply the sign function as an alternative to the unit step function.
And the decision tree \ref{Fig.main1} can be reorganized as following 
\begin{equation}\label{Tree}
   T(\mathbf{x})=v[i], i=\arg\max(\tilde{\mathrm{B}}\operatorname{sgn}(\tilde{S}\tilde{x})) 
\end{equation}
where $\tilde{\mathrm{B}}=\operatorname{diag}(\|B_1\|_1,\cdots, \|B_L\|_1)^{-1}B$;
the augmented matrix $\tilde{S}=[S \,\,\vec{b}]\in\mathbb{R}^{m\times(n+1)}$;
the augmented vector $\tilde{x}=(x^t \,\, -1)^T\in\mathbb{R}^{n+1}$;
other the notations are same as before except
$$\mathrm{B}=\begin{pmatrix}
-1 & -1 & 0 & -1 & 0\\
-1 & -1 & 0 & +1 & -1\\
-1 & -1 & 0 & +1 & +1\\
-1 & +1 & 0 & 0 & 0\\
+1 & 0 & -1 & 0 & 0\\
+1 & 0 & +1 & 0 & 0\\
\end{pmatrix}.$$

There is a ternary code scheme to substitute for the bitvector matrices.
Based on the theorem \eqref{theorem_zhang}, each leaf corresponds to a ternary row vector called decision vector or devector for short
where the positive ones indicate the tests  its travelers must pass;
the  negative ones indicate the tests  its travelers must fail;
the zeroes indicate the tests its travelers are absent.
The number of non-zeroes is the depth of the leaf node to the root.

We call this \emph{pattern matching interpretation} of decision trees\eqref{GT}
because it is to select the prediction based on similarity between  decision vector of leaf node and the tests results of input samples.

\begin{table}  
  \caption{The Dictionary on Two Interpretation Schemes on  Decision Trees}  
  \begin{center}  
  \begin{tabular}{p{5cm}|p{5cm}}
  \hline  
  Logical Interpretation & Pattern Matching Interpretation \\
  \hline  
  \hline
  Based on the columns of the bitvector matrices & Based on the rows of the ternary pattern matrices \\
  \hline
  Equivalent to the logical `AND' & Equivalent to  the `decision path' \\
  \hline
  Bitvector-based & Pattern-based \\
  \hline
  Binarized ReLU/Unit step function &  Sign/signum function \\
  \hline
  Test function & Decision function \\
  \hline
  Test nodes & Decision nodes\\
  \hline
  Terminal/leaf nodes & Prediction nodes\\
  \hline
  \end{tabular}  
  \end{center}  
\end{table}

\subsubsection{The Index Finding Function}

Note that in \eqref{tree} the prediction phase can be regarded as an inner product 
$$v[i]=\left<v,  \vec 1_i\right>=\sum_{n=1}^{L}v[n] 1_i(n)$$
where $\vec 1_i=(0,0,\cdots,1,0,\cdots, 0)$
and $1_i(n)=1$ if $n=i$ otherwise $1_i(n)=0$.
The operator $\arg\max$ is equivalent to choose an one-hot vector,
and such a vector is only non-zero at the index of the selected leaf.
Here we do not specify the data type or any requirement on the value vector $v$.
The element $v[i]$ can be anything even function.
Usually it is numerical for regression and categorical/discrete for classification.

In order to make \eqref{tree} differentiable, we  apply $\operatorname{softmax}$ to approximate the one-hot vector:

\begin{equation}\label{GT2}
  T(\mathbf{x}) =\left<\operatorname{softmax}(\mathrm{B}\,\sigma(\mathrm{S}\mathbf{x}-b)), p(\mathbf{x})\right> 
       =\sum_{i=1}^{L}e_i p_i(\mathbf{x}).
\end{equation}
Here $(e_1,\cdots,e_l)^T=\operatorname{softmax}(\mathrm{B}\,\sigma(\mathrm{S}\mathbf{x}-b))$.
In another word we use the softmax function as an alternative to the  navigation function in \href{https://arxiv.org/abs/1511.04056}{Mohammad Norouzi et al}\cite{norouzi2015efficient}.

Note that the we prefer the smaller index of the maximums of $h=\mathrm{B}\,\sigma(\mathrm{S}\mathbf{x}-b)$, 
it is best to find better approximation of the one-hot vector as shown in \cite{jang2016categorical, Sigsoftmax1, martins2016softmax, niculae2017regularized,laha2018controllable, correia2019adaptively}.

However, all above functions are coordinate independent which does not reflect your preference on minima index of the target leaf node perfectly.
Suppose $\mathbf{z}\in\mathbb{R}_{+}^{L}$, we would like to find the function that indicate first index of the maximums in the vector $z$.
For example, $\mathbf{z}=\vec{1}\in\mathbb{R}^{L}$, it is supposed to return $$(1, 0,\cdots,0)^T\in\mathbb{R}_{+}^{L};$$
if $\mathbf{z}=(1,2,\cdots, L-1,L)^T$, it is supposed to return $$(0,0,\cdots,0,1)^T\in\mathbb{R}_{+}^{L};$$
if the last $m$ elements of $z$ are equal to $m$, it is supposed to return $$(0,0,\cdots,\underbrace{0,0,\cdots, 0,1}_{\text{m elements}});$$
if the $i$th element of $z$ is the unique maximum, it is supposed to return the one-hot vector $\vec{1}_{1}$.
Mixed with the positive decreasing function $w(i)$ such as $w(i)=\frac{1}{i}$, we define `index sensitive' sparsemax function: 
$$\operatorname{sparsemax}(\mathbf{z})=\arg\min_{\mathbf{p}\in \triangle_{L-1}}\|\mathbf{p}-\mathbf{z}\|_{w}^2+\|\mathbf{p}\|_1$$
where the norm $\|\mathbf{p}-\mathbf{z}\|_{w}^2=\sum_{i} w_i(p_i-z_i)^2$.
It is a variant of \href{https://www.stat.cmu.edu/~ryantibs/papers/genlasso.pdf}{generalized lasso}.

In this setting, it is really analogous to the shallow  neural network.
If the bitvector matrix is the gene of decision tree,
this setting is the revolutionary mutation and
numerical optimization  is genome editing to train these models.

All the operators are common in deep learning such as matrix multiplication, ReLU and softmax.
There are one affine layer, a nonlinear layer and a softmax layer.
In some sense, decision trees are special type neural networks.
We call those models `logical decision machines'.
This representation is `mixed-precision'
because only $\mathrm{B}$ is 0-1 valued \footnotemark
\footnotetext{In a vanilla binary decision tree, the selection matrix is also 0-1 valued. Here we focus on oblique decision trees.}
and other notations are real.

\subsubsection{The Prediction}

\href{https://www.ijcai.org/Proceedings/2019/0476.pdf}{Yu Shi, Jian Li and Zhize Li} use piecewise linear regression trees (PL Trees)
to enhance the accuracy and efficiency of GBDT.
In fact, the prediction functions $\mathbf{p}=(p_1,\cdots, p_{L})^T$ can be substituted for the constant output value vector $\mathbf{v}=(v_1, v_2,\cdots, v_L)$
associated with the leaves of the decision tree.

\begin{equation}\label{decision spline}
  T(\mathbf{x})=p_i(\mathbf{x}),
\end{equation}
where $i$ is the first index of maximums of $\mathrm{B}\,\sigma(\mathrm{S}\mathbf{x}-\mathbf{t})$,
 i.e., $i=\min\{j\mid v_j=\max(v), v=\mathrm{B}\,\sigma(\mathrm{S}\mathbf{x}-\mathbf{t})\}$; 
and $p_i:\mathbb{R}^n\mapsto\mathbb{O}$, $\mathbb{O}$ is the domain of ground truth.
other notations are the same as shown in \eqref{tree}.
The function $f_i$ can be any type function.  
These are similar to ``Multivariate Adaptive Regression Splines''.

The prediction functions are diverse, configurable and dependent on the task and data set.
And each prediction function only process partial data.
The formula\eqref{decision spline} inherently contains model selection. 
The component prediction function $p_i$ associated with the leaf $i$ rely on the arrived samples rather than the whole sample set, 
i.e.,  every leaf only solve a specific problem of the task specially assigned to it. 
This modification keeps the identification of the problem separate from the solution to the problem.

\subsection{Unification of Binary Decision Trees}

We revisit the generalization of decision trees.
\href{http://pages.stat.wisc.edu/~loh/treeprogs/guide/grapes.pdf}{Probal Chaudhuri,  Wen-Da Lo,  Wei-Yin Loh,  Ching-Ching Yang}
propose generalized regression trees
by blending tree-structured nonparametric regression and adaptive recursive partitioning with maximum likelihood estimation.
\href{https://epub.wu.ac.at/676/1/document.pdf}{Torsten Hothorn, Torsten and Hornik et al}\cite{hothorn2006unbiased} embed tree-structured regression models
into a well defined theory of conditional inference procedures.
\href{https://www.researchgate.net/publication/261660212_Maximum_Likelihood_Regression_Trees}{Su X, Wang M C, Fan J, et al.}\cite{su2004maximum}
propose a method of constructing regression trees within the framework of maximum likelihood.
\href{http://pages.stat.wisc.edu/~loh/treeprogs/guide/LohISI14.pdf}{Wei-Yin Loh}\cite{loh2014fifty}
surveys the developments and briefly reviews
the key ideas behind some of the modern major decision tree algorithms.
\href{https://www.researchgate.net/profile/Richard_Berk}{Richard A. Berk} overviews the decision tree algorithms
in the monograph \emph{Statistical Learning from A Regression Perspective}.
Recently, \href{https://arxiv.org/abs/1906.10179}{Schlosser L, Hothorn T, Zeileis A, et al} \cite{schlosser2019the}
unify three popular  unbiased recursive partitioning approaches.
These generalization focus on the statistical inference.

In computer science community, hybrid decision trees integrate decision trees with other algorithms such as
\cite{DBLP:journals/jifs/Kotsiantis14, zhou2002hybrid, carvalho2004a}.
\href{https://link.springer.com/chapter/10.1007/978-981-10-1678-3_8}{Archana Panhalkar and Dharmpal Doye}
elaborate the various approaches of converting decision tree to hybridized decision tree.

Based on the concept of bitvector matrix $\mathrm{B}$, we formulate diverse decision trees in a unified way 

\begin{equation}\label{hybrid tree}
  T(\mathbf{x})=\left< p(\mathbf{x}), \sigma(\mathbf{h}) \right>, \mathbf{h}=\mathrm{B}\,\tau(f(\mathbf{x}))
\end{equation}
where $i$ is the first index of maximums of the hidden state $\mathbf{h}=\tau(f(\mathbf{x}))$;
$p_i(\mathbf{x})$ is the prediction function associated with the leaf node $i$ and $\mathbf{p}=(p_1(\mathbf{x}),\cdots, p_L(\mathbf{x}))$.

It is end-to-end. The feature is improved via the prediction evaluation.
If the nonlinear function $f$ is generalized decision tree, 
it is not hard to find yet another ensemble method of decision trees: composite decision trees or stacked decision trees.
In contrast to the additive boosted decision tree, stacked decision tree is a ``multiplicative model''.
This is another story on how to ensemble decision trees.

\section{Conclusion and Discussion}

We unify diverse decision trees or recursive partitioning methods in our novel parameterized framework of decision trees 
so that we can use the deep neural networks packages to deploy and optimize the different decision trees.
In theory, it is proved that a decision trees is determined by the tuple $(v, \mathrm{B}, \mathrm{S}, \mathbf{t})$ consisting of 2 matrices and 2 vectors.
Specially, we find some mathematical properties of the structure matrix $\mathrm{B}$.
We present how to unify some extension of decision trees in our new parameterization framework.
It is next step to search more efficient methods to jointly optimize the parameters of decision trees.
And we will study some theoretical issues in future such as the complexity of our novel representation and the universal approximation of decision trees

It is a regret that we do not cover the probabilistic decision tree.
\bibliographystyle{plain}
\bibliography{DM}

\end{document}